\documentclass[11pt]{article}

\usepackage[utf8]{inputenc}
\usepackage[T1]{fontenc}
\usepackage{lmodern}
\usepackage[a4paper,margin=1in]{geometry}
\usepackage{microtype}
\usepackage{graphicx}
\usepackage{multirow}
\usepackage{amsmath,amssymb,amsfonts}
\usepackage{amsthm}
\usepackage{mathrsfs}
\usepackage[title]{appendix}
\usepackage{xcolor}
\usepackage{textcomp}
\usepackage{booktabs}
\usepackage{algorithm}
\usepackage{algorithmicx}
\usepackage{algpseudocode}
\usepackage{listings}
\usepackage{longtable}
\usepackage{makecell}
\usepackage{pifont}
\usepackage{subfigure}
\usepackage{tcolorbox}
\tcbuselibrary{listings,breakable}
\usepackage[framemethod=tikz]{mdframed}
\usepackage{changepage}
\usepackage{url}
\usepackage[numbers,sort&compress]{natbib}
\usepackage{hyperref}
\hypersetup{
    colorlinks=true,
    linkcolor=blue,
    citecolor=blue,
    urlcolor=blue
}

\definecolor{lightgrey}{RGB}{235,235,235}
\definecolor{lightred}{RGB}{255,204,203}
\definecolor{lightblue}{RGB}{200,230,255}
\definecolor{lightyellow}{RGB}{255,255,224}
\definecolor{lightgreen}{RGB}{215,238,145}
\definecolor{lightpurple}{RGB}{230,230,250}
\definecolor{dkgreen}{RGB}{0,128,0}

\title{Explainable Detection of Depression Status Shifts from User Digital Traces}

\author{
Loris Belcastro\\
\texttt{loris.belcastro@unical.it}
\and
Francesco Gervino\\
\texttt{francesco.gervino@dimes.unical.it}
\and
Fabrizio Marozzo\\
\texttt{fmarozzo@dimes.unical.it}
\and
Domenico Talia\\
\texttt{domenico.talia@unical.it}
\and
Paolo Trunfio\\
\texttt{paolo.trunfio@unical.it}
}

\date{DIMES, University of Calabria, Ponte P. Bucci 41/C, 87036 Rende, Italy}

\begin{document}

\maketitle

\begingroup
\renewcommand{\thefootnote}{}
\footnotetext{This is a preprint of a manuscript submitted to Social Network Analysis and Mining.}
\addtocounter{footnote}{-1}
\endgroup

\begin{abstract}
Every day, users generate digital traces (e.g., social media posts, chats, and online interactions) that are inherently timestamped and may reflect aspects of their mental state. These traces can be organized into temporal trajectories that capture how a user's mental health signals evolve, including phases of improvement, deterioration, or stability.
In this work, we propose an explainable framework for detecting and analyzing depression-related status shifts in user digital traces. The approach combines multiple BERT-based models to extract complementary signals across different dimensions (e.g., sentiment, emotion, and depression severity). Such signals are then aggregated over time to construct user-level trajectories that are analyzed to identify meaningful change points.
To enhance interpretability, the framework integrates a large language model to generate concise and human-readable reports that describe the evolution of mental-health signals and highlight key transitions. 
We evaluate the framework on two social media datasets. Results show that the approach produces more coherent and informative summaries than direct LLM-based reporting, achieving higher coverage of user history, stronger temporal coherence, and improved sensitivity to change points. An ablation study confirms the contribution of each component, particularly temporal modeling and segmentation. Overall, the method provides an interpretable view of mental health signals over time, supporting research and decision making without aiming at clinical diagnosis.
\end{abstract}


\section{Introduction}\label{sec:intro}

In recent years, everyday digital traces, including private communications, web searches, mobility traces, and social media interactions, have increasingly become a medium through which individuals express emotions, opinions, behaviors, and personal experiences~\cite{demoor2023digitaltrace,wang2018digitaltraces,mobility-instagram-2017}. When analyzed longitudinally, these traces may implicitly reflect aspects of a person’s psychological condition and reveal how mental well-being evolves over time. Rather than providing a static representation, user-generated content often forms a temporal trajectory in which phases of deterioration, recovery, or relative stability can emerge through linguistic and behavioral patterns.

Several studies have investigated the use of social media data to automatically detect indicators of mental distress, including depression, major depressive disorder, and suicidal ideation, through natural language processing, machine learning, and deep learning techniques \citep{10.1145/3422824,zhang2022natural,tsugawa2015recognizing,Tadesse2019DetectionOD}. Early work showed that linguistic and behavioral signals extracted from platforms such as Twitter and Reddit can support the recognition of depressive symptoms, while more recent approaches have exploited transformer-based models and richer textual representations to improve classification performance \citep{poswiata-perelkiewicz-2022-opi,kerasiotis2024depression}. However, much of this literature still relies on post-level or user-level classification settings, whereas comparatively less attention has been devoted to modeling how mental-health signals evolve over time. Recent work on temporal boundaries in depression classification suggests that longitudinal segmentation can provide useful information for distinguishing depressive and control users \citep{villaperez2026comparative}.

Transformer-based architectures such as BERT and related language models have significantly advanced textual analysis by capturing contextual and semantic relationships within language. These models have achieved state-of-the-art performance in tasks including sentiment analysis, emotion recognition, and mental health classification \citep{kerasiotis2024depression,hridoy2024leveraging}. However, they are still commonly applied to classify posts independently, without explicitly modeling how mental health signals evolve across sequences of interactions over time. As a consequence, current approaches often provide limited insight into the temporal dynamics underlying changes in a user’s psychological condition and may fail to identify when meaningful shifts in mental state occur.

To address this limitation, this work introduces an explainable framework for tracing mental-health trajectories in individual digital traces. Our approach analyzes not only single posts but the entire sequence of posts associated with a user, modeling how linguistic, emotional, and psychological patterns evolve over time. The framework is based on a multi-classification strategy that combines several fine-tuned BERT-based models, each specialized in a specific dimension (e.g., sentiment, emotion, mental-health category) \citep{multi-class-LLMs-AIAI2024}. By integrating these complementary signals, the system builds a multidimensional representation of each post. A temporal reasoning module then processes the sequence of annotated posts to infer the user’s trajectory, explicitly identifying change points where the trajectory significantly changes direction (toward improvement or deterioration), as well as intervals of stability.

Explainability is a central part of the proposed framework, as interpretability is widely recognized as essential for the adoption of AI-based mental-health detection systems in clinical and high-stakes settings~\cite{joyce2023explainable,symanto2023xai}. Each BERT-based model produces signals that capture linguistic, emotional, and semantic patterns associated with the user’s mental state. These signals are aggregated over time to provide a global view of how such patterns evolve across the user’s timeline. To make this information accessible to clinicians and researchers, the framework employs a large language model (LLM) to transform the enriched signals into concise and human-readable summaries. These summaries describe the overall trend and highlight key transitions, linking them to recurring expressions, themes, and behavioral patterns observed in the data. More broadly, this type of system could be integrated into tools that support users and professionals (e.g., clinicians, psychologists, researchers) by highlighting meaningful changes in mental-health–related signals and facilitating reflection and early awareness. 
Unlike traditional explainable AI approaches, which primarily focus on feature attribution at the token level, our framework provides \emph{trajectory-level explainability}, where explanations are expressed as structured summaries of temporal patterns, phases, and transitions that directly reflect the evolution of mental-health signals over time.

In this work, we apply and evaluate our framework on two social media datasets from Reddit and Twitter, focusing on users who utilize online platforms as a personal space to share experiences. This enables the construction of detailed timelines that capture the evolution of their behavior and emotions over time. Experimental results show that our approach, which explicitly models temporal dynamics, is more effective at detecting meaningful shifts in mental-health signals compared to methods that ignore time or use a single model. In addition, the framework preserves interpretability by producing structured and human-readable summaries.
The system is not intended for clinical diagnosis; rather, it is designed as a decision-support and research tool that provides transparent signals and interpretable summaries. These outputs must be carefully interpreted and validated by domain experts, such as clinicians or researchers, within an appropriate clinical and ethical context.

The remainder of this paper is organized as follows. Section~\ref{sec:related} reviews existing literature on mental health detection from social data. Section~\ref{sec:methodology} presents the proposed framework. Section~\ref{sec:experiments} reports the experimental evaluation and comparative analysis. Section~\ref{sec:discussion} discusses limitations and future research directions. Finally, Section~\ref{sec:conclusions} concludes the paper.

\section{Related work}
\label{sec:related}

Digital traces generated through online interactions have increasingly been used to investigate indicators of mental health conditions, including depression, anxiety, suicidal ideation, and eating disorders. Several studies have shown that linguistic and behavioral patterns extracted from social media platforms can provide meaningful signals associated with psychological distress \citep{10.1145/3422824,zhang2022natural,li2018detecting,LYONS2018207}. With the increasing availability of user-generated data from platforms such as Twitter, Reddit, and Facebook, machine learning and NLP techniques have been widely adopted to automatically identify signals associated with mental distress \citep{tsugawa2015recognizing,10.1145/2531602.2531675,Tadesse2019DetectionOD,10108975}.

Early approaches mainly relied on traditional machine learning algorithms, including Naive Bayes, Random Forests, support vector machines, and handcrafted linguistic or affective features \citep{alsagri2020machine,peng-dep-2019,CHIONG2021104499,islam2018}. Several studies demonstrated that lexical, emotional, and behavioral signals extracted from social media activity can provide meaningful indicators of depressive symptoms and psychological vulnerability \citep{shen2017depression,orabi2018deep,stephen2019detecting}.

The introduction of transformer architectures, particularly BERT and its variants, significantly advanced the field by enabling richer contextual and semantic representations of user-generated text \citep{devlin2018bert,liu2019roberta,yang2019xlnet}. Recent studies have shown that transformer-based models can effectively capture subtle linguistic patterns associated with depression and mental distress \citep{poswiata-perelkiewicz-2022-opi,kerasiotis2024depression,kanahuati2024depressive,hridoy2024leveraging}. More recent research has also explored the integration of large language models (LLMs) and explainable artificial intelligence (XAI) techniques to improve the interpretability of mental health prediction systems \citep{zogan2022explainable,hameed2025xai,almasud2025deepxai,belcastro2025bertxdd}. In particular, approaches based on SHAP, LIME, and natural-language explanations have been proposed to bridge the gap between model predictions and human-understandable psychological interpretations \citep{lundberg2017unified,ribeiro2016should,guidotti2018survey,sahoo2025limebert,NAJM2026867}.

Beyond post-level classification, several recent studies have emphasized the importance of modeling temporal dynamics and longitudinal behavioral patterns. Villa-P{\'e}rez et al.~\citep{villaperez2026comparative} demonstrated that temporal segmentation of user timelines can improve depression classification performance by distinguishing behavioral patterns before and after self-reported diagnoses. Other works have investigated sequential and trajectory-aware approaches to reconstruct the evolution of depressive symptoms over time using recurrent neural networks, hybrid architectures, and transformer-based representations \citep{spiliotis2025comparative,tejaswini2025hybrid,ganji2025hybrid}. In parallel, topic modeling and LLM-driven semantic analysis have increasingly been used to identify thematic transitions and evolving psychological trends within social media discussions \citep{zhao2025topicllm,liu2025twitter,thakur2025topicmodel}.

Despite these advances, most existing approaches still focus on isolated tasks such as classification, topic extraction, or explanation independently. Comparatively less attention has been devoted to unified frameworks capable of integrating heterogeneous digital traces, temporal trajectory modeling, explainable classification, and interactive narrative interpretation within a single end-to-end pipeline. Moreover, many existing systems provide static predictions without explicitly modeling how psychological states evolve over time or how users and experts can interactively explore such trajectories.

To address these limitations, the framework proposed in this work combines: $(i)$ heterogeneous digital trace acquisition and preprocessing, $(ii)$ BERT-based multi-dimensional classification for extracting mental-health–related signals, $(iii)$ temporal trajectory construction with change-point detection to model behavioral evolution over time, and $(iv)$ an LLM-driven reporting layer for generating human-interpretable outputs. Unlike prior work focused on isolated prediction tasks, the proposed approach integrates semantic enrichment, temporal analysis, explainability, and interactive exploration within a unified framework. The framework can be extended with a Retrieval-Augmented Generation (RAG) mechanism, allowing users and domain experts to query trajectories, inspect contextual information associated with specific periods, and obtain narrative-level explanations of behavioral changes over time.

\section{Proposed methodology}\label{sec:methodology}

\begin{figure}[htb!] 
\centering \includegraphics[width=1\linewidth]{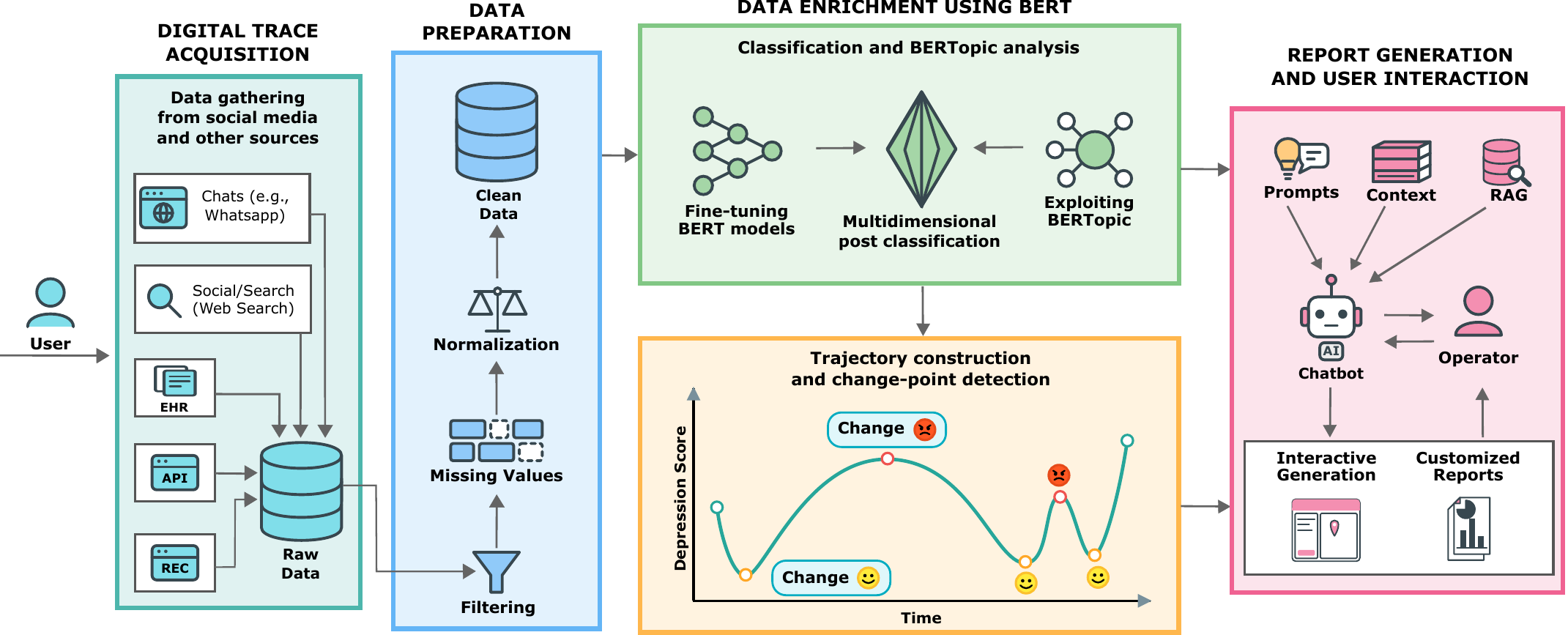} \caption{Execution flow of the proposed framework.} \label{fig:proposed_framework} 
\end{figure}

The proposed framework represents mental-health–related trajectories through a structured pipeline that integrates heterogeneous digital traces, semantic enrichment via BERT-based models, and temporal reasoning, as illustrated in Figure~\ref{fig:proposed_framework}. The workflow is organized into five main stages: $(i)$ digital trace acquisition, $(ii)$ data preparation, $(iii)$ multidimensional data enrichment using BERT, $(iv)$ trajectory construction and change-point detection, and $(v)$ report generation with user interaction.

\subsection{Digital trace acquisition}

The framework is designed to ingest heterogeneous textual data generated by users across multiple digital environments, such as social media posts, chats, web searches, transcripts, clinical records, and data obtained through APIs or digital services. In this study, we evaluate the framework on social media traces, while other sources represent possible extensions of the same event-based representation. Each piece of information is treated as a \emph{digital trace}, potentially encoding implicit signals about the mental state of the user.
To keep the framework general, all traces are mapped into a unified event representation. Each \emph{event} is defined as a tuple that contains a user identifier, a timestamp, a source label, and the text content. This representation lets the framework work with data from different sources while preserving the time order. For each user $u$, all events are merged and sorted in chronological order to form a single timeline $\{e_{u,1}, \dots, e_{u,n}\}$, where each event can be written as $e_{u,i} = (u, d_{u,i}, \rho_{u,i}, t_{u,i})$, with timestamp $d_{u,i}$, source label $\rho_{u,i}$, and text content $t_{u,i}$.

\subsection{Data preparation}

The data preparation stage ensures that heterogeneous and potentially noisy traces are transformed into a consistent and reliable format suitable for downstream analysis. This process includes filtering, missing value handling, and normalization. Filtering removes duplicated, corrupted, or non-informative events, as well as source-specific irrelevant content. Missing values are handled by either inferring attributes (e.g., estimating timestamps from neighboring events) or discarding incomplete records when reconstruction is not reliable. Finally, normalization standardizes textual content (e.g., lowercasing and cleaning) and harmonizes timestamp formats across sources, thereby enabling robust inputs for subsequent BERT-based analysis.

\subsection{Multidimensional data classification using BERT}

To extract depression-related signals from each event, the framework employs BERT-based models trained to classify data across several dimensions, such as sentiment (e.g., positive, neutral, negative), emotion (e.g., sadness, anger, fear, joy), and discrete levels of depression severity (no, moderate, severe). In addition to the predicted label, the BERT-based classifier outputs a probability distribution over the set of classes:
\[
\boldsymbol{\pi}_{u,i} = \big(\pi_{u,i}^{(1)}, \pi_{u,i}^{(2)}, \dots, \pi_{u,i}^{(C)}\big),
\quad \sum_{c=1}^{C} \pi_{u,i}^{(c)} = 1,
\]
where $\pi_{u,i}^{(c)}$ represents the probability that event $i$ of user $u$ belongs to class $c$ (e.g., positive, negative, or neutral for the sentiment classifier). 

In addition, BERTopic is employed to identify the main topics discussed in the dataset, providing a high-level understanding of thematic trends and emerging issues. This combined classification and topic modeling process enables a more comprehensive characterization of the data, supporting the identification of recurring themes, behavioral patterns, and depression-related signals by grouping semantically similar events. Although the current implementation is limited to English-language data, the proposed methodology can be extended to other languages by leveraging multilingual models such as mBERT, which is trained on several languages and provides strong cross-lingual capabilities. 

\subsection{Trajectory construction and change-point detection}

Once each event has been annotated, the outputs of the depression classifier are aggregated at the daily level to compute an overall depression severity score for each day in the user's timeline. Let $\{d_{u,1}, \dots, d_{u,n}\}$ denote the ordered set of days on which user $u$ is active, and $\mathcal{C} = \{\text{no},\, \text{moderate},\, \text{severe}\}$ the set of depression severity classes. For each event $i$, the classifier produces a probability distribution over $\mathcal{C}$, which is mapped to a scalar score as:
\[
s_{u,i} = \sum_{c \in \mathcal{C}} w_c \, \pi_{u,i}^{(c)},
\]
where $\pi_{u,i}^{(c)}$ denotes the probability assigned to class $c$, and $w_c$ are predefined weights reflecting increasing severity levels (e.g., $w_{\text{no}}=0$, $w_{\text{moderate}}=1$, and $w_{\text{severe}}=2$). A higher score value indicates a higher level of depression. For each day $d_{u,j}$, we compute a daily score by aggregating all events occurring on that day:
\[
r_{u,j} = \frac{1}{|\mathcal{I}_{u,j}|} \sum_{i \in \mathcal{I}_{u,j}} s_{u,i},
\]
where $\mathcal{I}_{u,j}$ is the set of events associated with user $u$ on day $d_{u,j}$. The resulting sequence $\mathbf{r}_u = \{r_{u,1}, \dots, r_{u,n}\}$ defines a univariate time series representing the user's estimated depressive state over time. To reduce noise and highlight long-term trends, the trajectory is smoothed using a moving average filter, yielding the sequence $\tilde{\mathbf{r}}_u = \{\tilde{r}_{u,1}, \dots, \tilde{r}_{u,n}\}$, which represents the user’s depression status trajectory over time.

\begin{algorithm}[t]
\caption{Top-down piecewise linear segmentation}
\label{alg:topdown}
{\fontsize{9}{10.5}\selectfont
\begin{algorithmic}[1]

\Require Daily scores $\{(d_1,r_1),\dots,(d_n,r_n)\}$ sorted by date, desired number of segments $K$
\Ensure Ordered segments $\mathcal{S}$
\State Smooth the daily score series to obtain $\tilde{r}_1,\dots,\tilde{r}_n$
\State Define $x_i$ as the number of days elapsed from $d_1$ to $d_i$, and $y_i \gets \tilde{r}_i$
\State $\mathcal{S} \gets \{[1,n]\}$

\While{$|\mathcal{S}| < K$}
    \State $(e^\star, j^\star, p^\star) \gets (0,\mathrm{None},\mathrm{None})$
    \For{each segment $[a,b]$ in $\mathcal{S}$ at position $p$}
        \If{$b-a > 1$}
            \State $j \gets \arg\max_{a<i<b} \mathrm{dist}\big((x_i,y_i), \ell_{a,b}\big)$
            \State $e \gets \mathrm{dist}\big((x_j,y_j), \ell_{a,b}\big)$
            \If{$e > e^\star$}
                \State $(e^\star, j^\star, p^\star) \gets (e,j,p)$
            \EndIf
        \EndIf
    \EndFor
    \If{$j^\star = \mathrm{None}$ or $e^\star \le 0$}
        \State \textbf{break}
    \EndIf
    \State Let $[a,b]$ be the segment in $\mathcal{S}$ at position $p^\star$
    \State Replace $[a,b]$ with $[a,j^\star]$ and $[j^\star,b]$
\EndWhile

\State Sort $\mathcal{S}$ by starting index
\State \Return $\mathcal{S}$

\end{algorithmic}
}
\end{algorithm}

On the smoothed trajectory, we apply a top-down piecewise linear segmentation procedure to identify a small number of change points, as detailed in Algorithm~\ref{alg:topdown}. This procedure is inspired by the Ramer--Douglas--Peucker algorithm for polygonal curve approximation~\cite{ramer1972iterative,douglas1973algorithms}, but it is adapted here to time-series segmentation by selecting a fixed maximum number of segments $K$ rather than using a distance-tolerance stopping criterion. Let $\{(d_1,r_1), \dots, (d_n,r_n)\}$ denote the daily scores sorted by date, and let $\tilde{r}_1,\dots,\tilde{r}_n$ be the corresponding smoothed values obtained with a centered moving average. For each day $d_i$, we define $x_i$ as the number of days elapsed from the first observation $d_1$, and we set $y_i = \tilde{r}_i$. A segment is then defined by an interval of indices $[a,b]$, with $1 \leq a < b \leq n$, and is approximated by the straight line $\ell_{a,b}$ joining the two endpoints $(x_a,y_a)$ and $(x_b,y_b)$.

For each segment $[a,b]$, we measure its approximation error as the maximum perpendicular distance between the line $\ell_{a,b}$ and the intermediate points:
\[
E(a,b) = \max_{a < j < b} \mathrm{dist}\big((x_j,y_j), \ell_{a,b}\big).
\]
This quantity captures how well a single straight line approximates the trajectory within the interval. A large value of $E(a,b)$ indicates that the segment contains an internal change in trend that is not well represented by a linear approximation.

The procedure starts from a single segment covering the entire trajectory, i.e., $\mathcal{S} = \{[1,n]\}$. At each iteration, the approximation error is evaluated for all current segments. Segments with fewer than three points, i.e., such that $b-a \leq 1$, are not further split, since they contain no internal point on which the deviation can be evaluated. Among all current segments, the one with the largest approximation error is selected. The split position $j^\star$ is then identified as the internal point with maximum distance from the line joining the segment endpoints, and the selected segment is replaced by the two subsegments $[a,j^\star]$ and $[j^\star,b]$.

This process is repeated until the desired number of segments $K$ is reached or no segment can be further split with positive approximation error. After termination, the segments are sorted by their starting index, yielding an ordered segmentation
\[
[a_{u,1}, b_{u,1}], \dots, [a_{u,K_u}, b_{u,K_u}],
\]
where $K_u \leq K$. The internal boundaries of consecutive segments can be interpreted as change points of the trajectory. Each resulting segment therefore represents a coherent temporal interval that is subsequently analyzed and summarized by the LLM-based reporting component. In the experiments, we set $K=10$ to obtain a compact yet sufficiently detailed phase representation; sensitivity to this choice is indirectly assessed through the fixed-window and no-segmentation ablation variants.

\subsection{Report generation and user interaction}

The final step of the proposed framework transforms the segmented trajectory into a concise and human-readable report. After segmentation, each segment $[a_{u,k}, b_{u,k}]$ is provided to a generative AI model together with the events belonging to that time interval and a small set of trajectory descriptors, such as the average severity level, the local trend, and the segment position in the overall timeline. To improve report quality, this information can be enriched with additional contextual evidence, such as operator annotations, relevant life events, or background notes that help interpret the trajectory. The framework can also incorporate a RAG component when the LLM needs access to evidence that should not be compressed into the trajectory alone. For example, the retriever may supply original data explaining a worsening phase, metadata about a specific interval, previously identified key events, clinician notes stored in the system, or domain-specific reference material. 
The report generation process is explicitly grounded in the structured signals produced by the previous stages of the pipeline. In particular, each segment is associated with quantitative descriptors (e.g., average severity, trend direction, and temporal position) and a well-defined subset of events. The LLM is guided to generate explanations conditioned on this structured representation, through prompts that explicitly encode segment-level descriptors and associated events. This design encourages the model to produce summaries that reflect the underlying temporal dynamics rather than arbitrary narrative generation.

Starting from these inputs, the LLM generates a short natural-language report describing the inferred phase. These reports are designed to support clinicians and researchers by summarizing how depression-related signals evolve over time and by highlighting phases of worsening or improvement. In this way, the report does not replace the underlying trajectory, but presents it in a form that is easier to read, inspect, discuss, and validate.

Once the report has been generated, the operator can interact with the LLM to better understand specific parts of the trajectory. For example, the operator may ask what changed around a certain date, request the evidence behind a worsening phase, or ask for more details about a specific segment. In these cases, the RAG component can support the response by retrieving the most relevant information, such as original posts, segment summaries, trajectory statistics, clinician notes, or other useful external material. This allows the LLM to provide answers that are more accurate and better grounded in the available evidence.

\section{Experimental Results} 
\label{sec:experiments}

This section evaluates the proposed framework from both a qualitative and a quantitative perspective. We first describe the datasets used in the paper and motivate their suitability for longitudinal trajectory analysis. We then illustrate, through a small set of representative users, how raw writings are transformed into daily depression scores, segmented trajectories, and final narrative reports. Finally, we compare direct LLM-based summarization against the proposed trajectory-aware strategy and analyze the contribution of the main pipeline components through an ablation study.

To assess the effectiveness of our framework, we used two datasets: $(i)$ \textit{eRisk 2018}~\cite{losadaetal2018}, a widely adopted benchmark for the early detection of depression from Reddit posts, and $(ii)$ the \emph{Mental Health Social Media} dataset available on Kaggle\footnote{\url{https://www.kaggle.com/datasets/infamouscoder/mental-health-social-media}}. The eRisk collection is organized at the \emph{user level}, where each subject is associated with a chronologically ordered history of writings, including both posts and comments. The official release contains 1,707 users and 1,076,582 posts. The \emph{Mental Health Social Media} dataset contains 20,000 Twitter posts annotated with mental-health-related labels and spans a broader range of mental-health conditions, including depression. For the Mental Health Social Media dataset, trajectory-based analysis is applied only to records for which temporal information is available.
To support reproducibility, the implementation of the proposed framework is made publicly available\footnote{\url{https://github.com/SCAlabUnical/X-MiND}}. Due to the sensitivity of mental-health-related user data, we provide instructions for obtaining the original datasets from their official sources rather than redistributing raw posts.

\subsection{Qualitative analysis of representative users}
\label{sec:dataset}

\begin{figure}[!t]
    \centering
    
    \subfigure[User 2714.\label{fig:traj_2714}]{
        \includegraphics[width=0.45\linewidth]{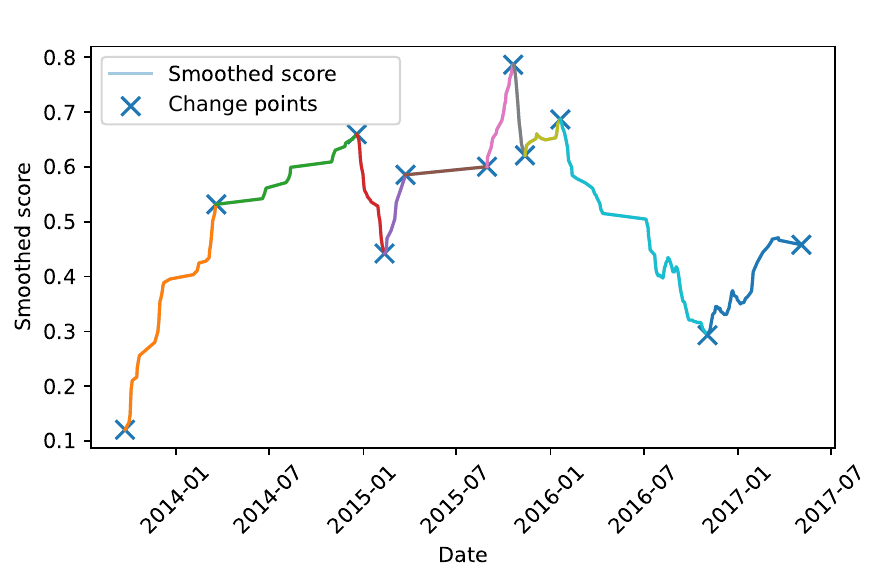}
    }
    \hspace{0.02cm}
     \subfigure[User 9280.\label{fig:traj_9280}]{
        \includegraphics[width=0.45\linewidth]{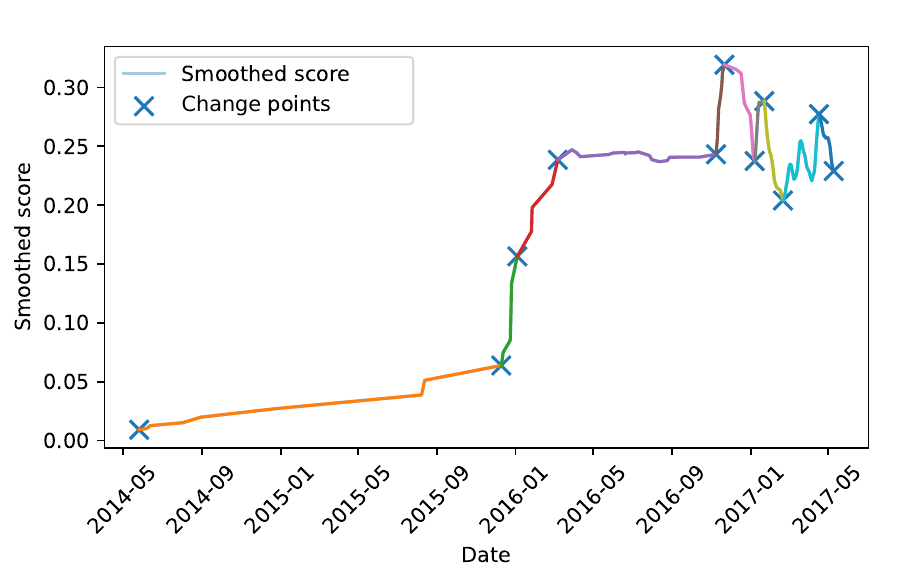}
    }

    \vspace{0.2cm}
    \subfigure[User 1257.\label{fig:traj_1257}]{
        \includegraphics[width=0.45\linewidth]{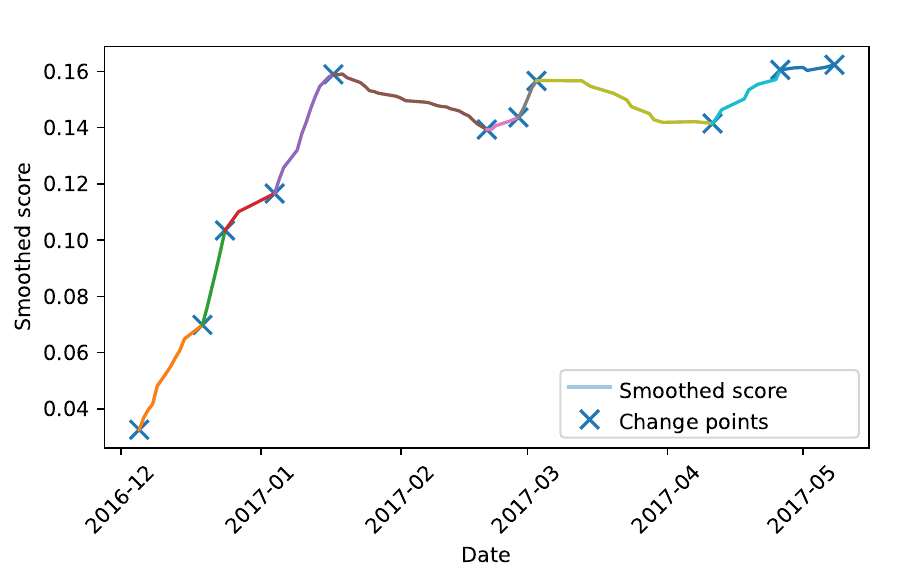}
    }    
    \hspace{0.02cm}
   \subfigure[User 3307.\label{fig:traj_3307}]{
        \includegraphics[width=0.45\linewidth]{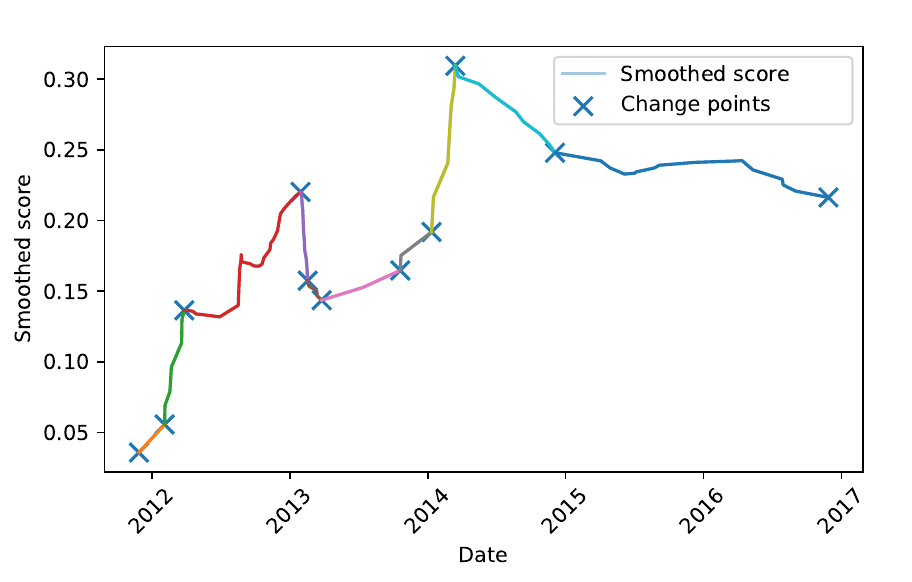}
    }
    
    \caption{Examples of depression trajectories and their piecewise linear segmentations ($K=10$) for four representative users. Each panel shows the smoothed daily depression scores together with the segments identified by the top-down algorithm and the corresponding change points.}
    \label{fig:multiuser_trajectories}
\end{figure}

To clarify how the proposed framework operates, we analyze four representative users from the eRisk 2018 dataset, namely IDs 1257, 2714, 3307, and 9280, selected because they exhibit different temporal profiles. As an example, Table~\ref{tab:user2714_examples} reports selected posts from User~2714, together with their dates, the depression label predicted by the classifier (\emph{no}, \emph{moderate}, \emph{severe}), and the corresponding aggregated daily score. Figure~\ref{fig:multiuser_trajectories} shows the smoothed trajectories and the corresponding piecewise linear segmentations for the four users. User~2714 presents a long, non-monotonic trajectory with several relapses, whereas User~9280 shows a more localized episode of deterioration surrounded by relatively stable intervals. User~1257 follows a smoother and more regular trajectory, with fewer abrupt transitions, while User~3307 is characterized by a pronounced initial phase followed by a long, low-severity plateau. Overall, the framework does not impose a fixed temporal model on all users; instead, it produces individualized, phase-based representations that capture sustained deterioration, gradual improvement, temporary relapses, and relatively stable conditions.

\begin{table}[!t]
\centering
\caption{Representative posts of user ID 2714 (eRisk 2018) with depression level and daily score.}
\label{tab:user2714_examples}
\fontsize{7pt}{8pt}\normalfont{
\begin{tabular}{llccc}
\hline
\textbf{Date} & \textbf{Post excerpt} &\textbf{\begin{tabular}[c]{@{}c@{}}Depression\\Level\end{tabular}} & \textbf{\begin{tabular}[c]{@{}c@{}}Class\\Probability\end{tabular}} & \textbf{\begin{tabular}[c]{@{}c@{}}Daily\\Score\end{tabular}} \\
\hline

\addlinespace[1.5ex]
\textit{2013-10-04} &
\textit{\begin{tabular}[c]{@{}l@{}}%
Well done, that's a great achievement!\\ Better than mine: 0.\\
B\dots or stopping! If you don't mind \\me asking, how did you stop?
\end{tabular}} &
No & 0.983 &
0.057 \\

\addlinespace[1.5ex]
\textit{2014-08-11} &
\textit{\begin{tabular}[c]{@{}l@{}}%
Hi, I've been diagnosed with PTSD,\\ Major Depressive Disorder and I hear voices.\\ I've been hospitalized once for 7 months.
\end{tabular}} &
Severe &
0.914 & 0.585\\

\addlinespace[1.5ex]
\textit{2015-10-19} &
\textit{\begin{tabular}[c]{@{}l@{}}%
My death is likely to send my mum back\\ to hospital.
The person \dots me and he has said \\once I kill myself,
he won't be far behind.
\end{tabular}} &
Moderate & 0.553 &
0.261 \\

\hline
\end{tabular}
}
\end{table}

\subsection{Report generation with generative models}
\label{sec:report}

Our framework aims at transforming the inferred trajectory into a concise and human-readable report. These reports are intended as decision-support outputs for clinicians and researchers: they should summarize the temporal evolution of depression-related signals, highlight phases of worsening or improvement, and provide a narrative that remains grounded in the user’s own language without making diagnostic claims. To this end, we compare two configurations. The first, denoted as \texttt{GPT-base}, is a direct summarization baseline. In this setting, all events of a user are concatenated in chronological order and passed to the LLM with a generic prompt to summarize the user’s mental-health history. The model is provided with only the raw texts and dates, and does not have access to the outputs of the proposed trajectory analysis.

The second configuration, denoted as \texttt{GPT-traj}, exploits the full pipeline described in Section~\ref{sec:methodology}. Posts are first enriched through the multidimensional classification, then aggregated into daily severity scores, smoothed, and segmented into phases by identifying the change points. Each phase is summarized separately by conditioning the LLM on the posts belonging to that interval together with basic segment-level descriptors such as average severity and trend direction. Then, a second LLM call composes these phase-level descriptions into a single global report, explicitly structured around the sequence of phases and the main change points.

To illustrate the qualitative difference between the two configurations, we consider User 2714, which is particularly suitable as an example because the timeline spans several years and contains both acute crisis periods and more stable intervals.
Below we report a shortened example of the type of summary produced using the \texttt{GPT-base} approach.

\begin{tcolorbox}[breakable=true, boxsep=0pt, left=3mm, right=3mm, top=2mm, bottom=2mm, sharp corners, boxrule=0pt, fontupper=\fontsize{9}{10}\selectfont, colback=lightgrey]
\textbf{\texttt{Baseline report for User 2714 (\texttt{GPT-base})}} 
\vspace{0.15cm}

Across several years, the user’s posts reveal a life narrative dominated by psychological distress, with periods of acute crisis and searching for ways to cope. From the earliest messages in 2013, the emotional tone is colored by anxiety and mistrust, as the user speaks about feeling watched by their college. This persecutory concern suggests heightened suspicion and a sense that their environment is unsafe. Early posts explicitly name mental health deterioration rather than fluctuation, establishing a baseline of distress. The user frequently expresses isolation and uncertainty about who to talk to, acknowledging suffering without a clear route toward care. Helplessness becomes a thematic thread suggesting persistent rather than situational distress.
\vspace{0.15cm}

Amidst these struggles, the user offers warmth and validation to others, highlighting an asymmetry where they recognize success in others more easily than in themselves. Their interest in metal music serves as a form of catharsis and remains a stable reference point, indicating their emotional life is not entirely flattened. Over time, the language reflects chronic mental health problems, difficulty functioning, and a desire to escape into nature. This wish to disappear into a forest is mixed with pessimistic resignation that such a refuge may be unattainable.
\vspace{0.15cm}

Later, the user explicitly asks how to learn to dissociate, signaling a concerning escalation from seeking community support to wishing for an internal mechanism to shut down awareness. This suggests they feel unable to bear their emotional state in full consciousness. The overall trajectory shifts from anxious vigilance and worsening symptoms to entrenched weariness and withdrawal. While the sense of persecution fades, the craving for psychological distance and ``off switches'' grows. However, the user’s continued engagement in the community indicates that, despite hopeless thoughts, an enduring drive to survive and adapt persists.
\end{tcolorbox}

This baseline summary captures the general tone of chronic depression, but it tends to merge the entire history into a single narrative arc. As a consequence, changes over time are described only loosely, and the distinction between temporally separate phases remains weak.

In the following, we present the corresponding report produced by \texttt{GPT-traj}, in which the LLM leverages the segmented trajectory and phase-level summaries.

\begin{tcolorbox}[breakable=true, boxsep=0pt, left=3mm, right=3mm, top=2mm, bottom=2mm, sharp corners, boxrule=0pt, fontupper=\fontsize{9}{10}\selectfont, colback=lightblue]
\textbf{\texttt{Trajectory-aware report for user 2714 (\texttt{GPT-traj})}} 
\vspace{0.15cm}

\textit{Phase 1 (from 2013-09-23 to 2014-12-19)}:
Persistent despair, paranoia, and recurrent crisis-related expressions characterize this period, with worsening anxiety and PTSD symptoms. Over time, distress intensifies alongside self-harm references and increasingly controlling voices. By late 2014, feelings of unreality, failure, and self-disgust dominate, alongside escalating depersonalization.
\vspace{0.15cm}

\textit{Phase 2 (from 2014-12-19 to 2015-03-24)}:
Mood is consistently bleak, marked by intense self-hatred and self-harm references framed as necessary for survival. The user moves from hopelessness to escalating crisis, with recurrent suicidal ideation, psychosis, PTSD triggers, and trauma-related distress repeatedly overwhelming coping attempts, despite brief positive comments.
\vspace{0.15cm}

\textit{Phase 3 (from 2015-03-24 to 2015-10-20)}:
Despair escalates into recurrent suicidal ideation and crisis-related expressions, with early posts framing notes as goodbyes. Frustration at surviving persists alongside intense psychotic distress and urges to self-harm. The trajectory trends toward desperation and resignation, despite intermittent therapy and medication.
\vspace{0.15cm}

\textit{Phase 4 (from 2015-10-20 to 2016-01-20)}:
Intense fear, paranoia, and psychosis drive repeated crisis-related expressions. The user feels ``used to'' suffering, intermittently sabotaging medication while trauma memories and nightmares intensify. Salient depressive posts involve acute distress and self-identification as a burden, maintaining a negative trajectory.
\vspace{0.15cm}

\textit{Phase 5 (from 2016-01-20 to 2017-05-04)}:
The emotional state remains bleak, dominated by exhaustion, withdrawal, and nihilistic engagement with morbid communities. Recurrent self-harm references and crisis-related expressions continue, with the user feeling ``no hope left''. Brief supportive moments are quickly overwhelmed by guilt and self-loathing.
\vspace{0.15cm}

\textit{Overall summary}:
The user’s emotional life is defined by persistent, often escalating despair and recurrent suicidal ideation that rarely recedes. Early anxiety and trauma distress fuse with psychotic experiences, voices, and depersonalization. Academic stress and family conflict repeatedly overwhelm fragile supports. Over time, the tone becomes entrenched, with the user feeling used to suffering and occasionally sabotaging treatment.
\end{tcolorbox}

Compared with \texttt{GPT-base}, the trajectory-aware report is explicitly organized into temporally ordered phases and makes the evolution of the user’s condition much more visible. Instead of collapsing the entire history into a generic summary of chronic distress, it connects specific themes and emotional patterns to well-defined intervals of time. This makes the resulting narrative more aligned with the underlying trajectory and more useful for understanding how depression-related signals develop across the user’s history.

\subsection{Aggregate evaluation}
\label{sec:aggregate_evaluation}

The qualitative examples discussed above suggest that trajectory-aware report generation yields richer and more temporally grounded summaries than direct LLM summarization. We now move to an aggregate evaluation in order to quantify this effect across a broader set of users.

Our aggregate analysis focuses on two complementary questions. First, does the trajectory-aware strategy cover the user’s history more completely and coherently than the baseline? Second, does the explicit use of segmentation and change points lead to better narrative reports from the point of view of independent judge models? To answer these questions, we evaluate report quality through two perspectives: topic coverage and \emph{LLM-as-a-judge} comparison.

\subsubsection{Topic Coverage}
\label{sec:coverage}

We consider the ability of the generated reports to capture the semantic content of the user's original timeline. In the baseline condition, the LLM processes the user's entire timeline as a single input, which tends to favor a high-level synthesis around dominant and recurring topics. In contrast, the trajectory-based approach requires a step-by-step analysis of the user's timeline, increasing the likelihood that temporally localized topics will be retained in the final report.

To quantitatively evaluate this aspect, for each user in the dataset, we extracted the top 15 topics from their posts using BERTopic. We then generated two reports for each user: one using the baseline approach (\texttt{GPT-base}) and one using the trajectory-based approach (\texttt{GPT-traj}). Finally, we used an external LLM (GPT 5.2) to evaluate how many of the extracted topics are actually covered in each report.
The results show a clear advantage for the trajectory-based approach. On average, \texttt{GPT-base} covers 40.62\% of the extracted topics, whereas \texttt{GPT-traj} covers 84.08\%. This substantial improvement indicates that the trajectory-based method provides significantly broader and more balanced topic coverage. Specifically, it is more effective at preserving not only the dominant depressive topics, but also secondary and stage-specific themes, which are often overlooked in the baseline setting. This aspect is particularly relevant in clinical and research contexts, where a more comprehensive representation of the user's experiences can support domain experts in interpreting longitudinal patterns without replacing professional assessment.

\begin{figure}[!t]
    \centering
    
    \subfigure[User 2714.\label{fig:topic_2714}]{
        \includegraphics[height=0.45\linewidth]{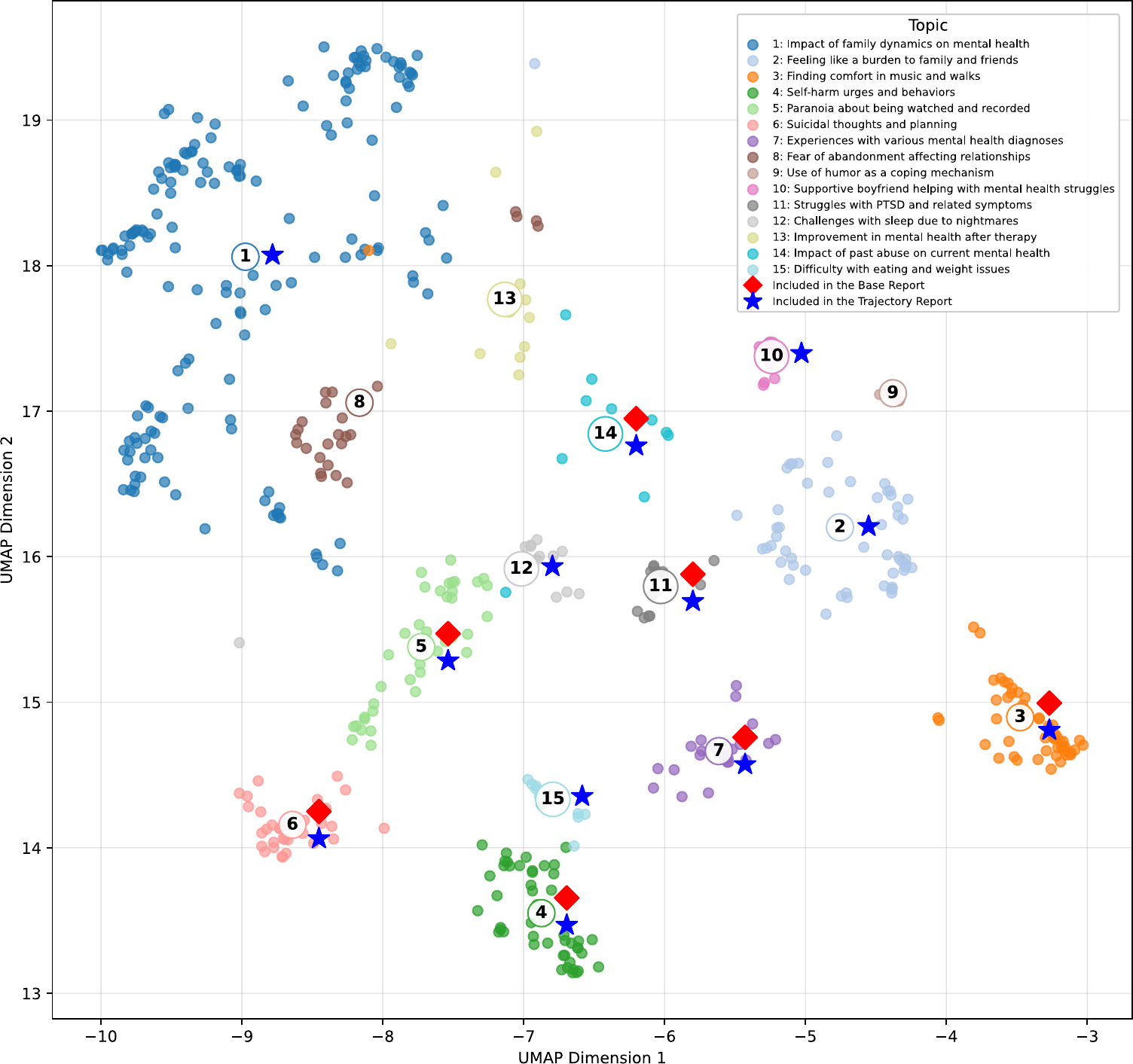}
    }
    \hspace{0.02cm}
    \subfigure[User 9280.\label{fig:topic_9280}]{
        \includegraphics[height=0.45\linewidth]{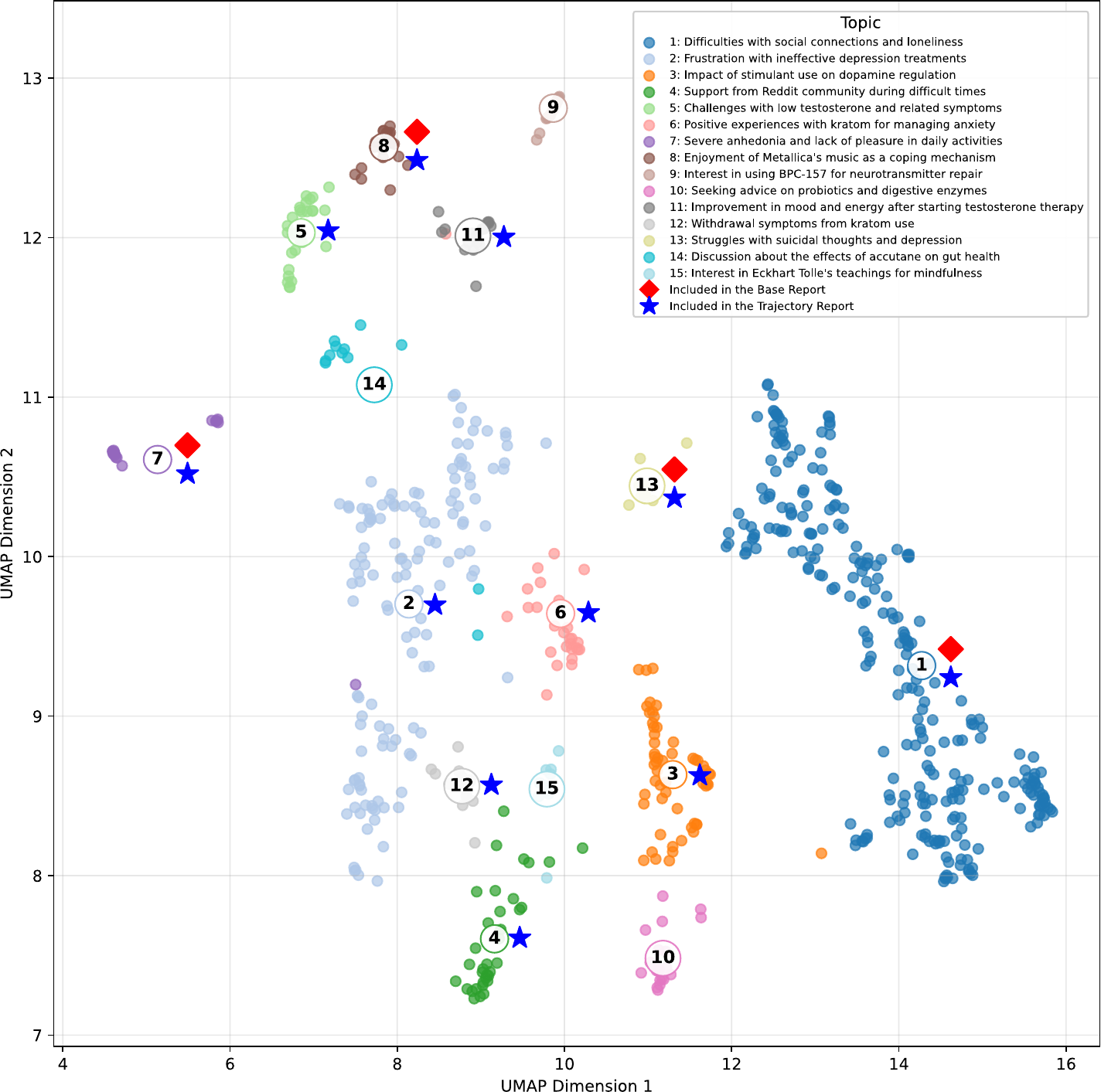}
    }
    
    \caption{Topic coverage for two representative users, showing the UMAP projection of the top 15 BERTopic topics extracted from each timeline; red diamonds indicate topics covered by \texttt{GPT-base}, while blue stars indicate topics covered by \texttt{GPT-traj}.}
    \label{fig:topic_coverage}
\end{figure}

Figure~\ref{fig:topic_coverage} illustrates this behavior for two representative users, namely User~2714 and User~9280. For User~2714, \texttt{GPT-base} covers 7 out of 15 topics, whereas \texttt{GPT-traj} covers 12 out of 15 topics. Similarly, for User~9280, \texttt{GPT-base} covers 4 out of 15 topics, while \texttt{GPT-traj} covers 11 out of 15 topics. These examples visually confirm that the trajectory-based method captures a wider portion of the user's topical space.

\subsubsection{LLM-as-a-judge}
\label{subsec:llm_judge}

For a more systematic evaluation, we adopt an \emph{LLM-as-a-judge} approach, where independent models compare, for each user, the reports produced by \texttt{GPT-base} and \texttt{GPT-traj}. For each user $u$ in the dataset, we generate two reports: $(i)$ a baseline report obtained by directly summarizing the full chronological post history, and $(ii)$ a trajectory-aware report derived from the multidimensional enrichment phase followed by segmented trajectory analysis. The two reports are presented in randomized order to four evaluator models: \emph{GPT 5.2}, \emph{Gemini 3.1 pro}, \emph{DeepSeek 3.2}, and \emph{Claude Opus 4.6}. Each model rates the reports on a five-point Likert scale (1 = lowest, 5 = highest) according to the following criteria:

\begin{itemize}
    \item \textit{Trajectory Coverage}: how well the report captures the main phases of the user’s history, rather than focusing on a limited portion of the timeline.
    \item \textit{Temporal Coherence}: how clearly the report describes changes over time and preserves a consistent chronology.
    \item \textit{Sensitivity to Change Points}: how effectively the report identifies and explains key transitions.
    \item \textit{Segment-Level Specificity}: the extent to which the report includes concrete, phase-specific details rather than generic statements.
    \item \textit{Overall Preference}: the overall usefulness and coherence of the report.
\end{itemize}

\begin{table}[!b]
\centering
\caption{Aggregated LLM-as-a-judge scores (1--5) comparing baseline (\texttt{Base}) and trajectory-based (\texttt{Traj}) report generation across different evaluator models and datasets. Higher values indicate better report quality.}
\label{tab:llm_judge_all}
\fontsize{10pt}{9pt}\selectfont
\resizebox{\textwidth}{!}{
\begin{tabular}{lcccccccccccccccc}
\toprule
\multirow{2}{*}{\textbf{Criterion}} 
& \multicolumn{8}{c}{\textbf{eRisk 2018 (Reddit)}} 
& \multicolumn{8}{c}{\textbf{Mental Health Social Media (Twitter)}} \\

\cmidrule(lr){2-9} \cmidrule(lr){10-17}
& \multicolumn{2}{c}{GPT}
& \multicolumn{2}{c}{Gemini}
& \multicolumn{2}{c}{DeepSeek}
& \multicolumn{2}{c}{Claude}
& \multicolumn{2}{c}{GPT}
& \multicolumn{2}{c}{Gemini}
& \multicolumn{2}{c}{DeepSeek}
& \multicolumn{2}{c}{Claude} \\

\cmidrule(lr){2-17}
& Base & Traj & Base & Traj & Base & Traj & Base & Traj
& Base & Traj & Base & Traj & Base & Traj & Base & Traj \\

\midrule

\makecell[l]{Trajectory\\coverage}
& 1.5 & 4.9 & 2.6 & 4.9 & 2.6 & 4.3 & 2.1 & 4.8
& 2.5 & 3.9 & 2.7 & 3.8 & 1.7& 3.3& 2.4 & 3.9 \\
\addlinespace[4pt]
\makecell[l]{Temporal\\coherence}
& 2.3 & 4.6 & 3.0 & 4.5 & 3.0 & 4.3 & 2.3 & 4.8
& 2.8 & 3.4 & 3.0 & 3.5 & 2.0& 3.6& 2.7 & 3.6 \\
\addlinespace[4pt]
\makecell[l]{Sensitivity to\\change points}
& 1.3 & 4.2 & 1.9 & 4.2 & 2.4 & 4.0 & 1.7 & 4.2
& 1.9& 3.2 & 2.2 & 3.3 & 1.3& 2.9& 2.1 & 3.4 \\
\addlinespace[4pt]
\makecell[l]{Segment-level\\specificity}
& 1.5 & 4.0 & 2.8 & 3.9 & 2.3 & 4.3 & 1.4 & 4.8
& 2.6 & 3.6 & 2.9 & 3.5 & 1.3& 3.7 & 2.4 & 3.8 \\
\addlinespace[4pt]
\makecell[l]{Overall\\preference}
& 1.5 & 4.1 & 2.5 & 4.0 & 2.5 & 3.9 & 1.9 & 4.8
& 2.6 & 3.5 & 2.6 & 3.4 & 1.8& 3.6& 2.4 & 3.6 \\

\bottomrule
\end{tabular}
}
\end{table}

Table~\ref{tab:llm_judge_all} shows and compares the average scores obtained with the four evaluation models. Overall, the \texttt{GPT-traj} approach is systematically preferred over \texttt{GPT-base}, producing reports with more coherent, time-contextualized, and informative narratives.
Trajectory-based reports achieve substantially greater coverage (e.g., 4.9 vs.\ 1.5 for GPT), capturing the entire depression-related signal evolution; in contrast, base reports often focus on limited portions of the timeline, omitting important phases. Regarding temporal coherence, trajectory-based outputs preserve the order of events and explicitly link observations over time (e.g., 4.6 vs.\ 2.3 for GPT), whereas base reports often distort the chronology or obscure significant changes.
The greatest improvement concerns sensitivity to change points (e.g., 4.2 vs.\ 1.3 for GPT): trajectory-based reports explicitly identify change points (e.g., emotional breakdowns, therapeutic changes, triggering events) and provide structured explanations of transitions; in contrast, base reports tend to rely on static and vague descriptions that overlook such discontinuities.
Improvements in segment-level specificity are also observed (e.g., 4.8 vs.\ 1.4 for Claude), with trajectory-based analyses incorporating more specific and detailed descriptions for each stage.

Similar results are observed on the Mental Health Social Media dataset, where trajectory-based reports consistently outperform the baseline across all evaluation criteria, although with slightly lower absolute scores due to the shorter and less structured nature of the posts.

\subsection{Ablation study}
\label{sec:ablation}

To assess the contribution of each component of the proposed framework, we evaluate several simplified variants of the full pipeline. 
The first variant, \texttt{NoSeg}, does not apply the dynamic segmentation defined in Algorithm~\ref{alg:topdown}; instead, the entire trajectory is computed and passed to the LLM as a whole, without any explicit phase decomposition. The second variant, \texttt{FixedWin}, replaces adaptive segmentation with a fixed partitioning into equal-length temporal windows (in our case, $K=10$). The third variant, \texttt{NoSmooth}, applies segmentation directly on raw daily scores without smoothing. The fourth variant, \texttt{NoStats}, keeps the segmented structure but removes numerical descriptors such as average severity and trend direction. We also include \texttt{Base} as the direct summarization baseline and \texttt{Full} as the complete framework.
All variants are evaluated using the same \emph{LLM-as-a-judge} protocol described in Section~\ref{subsec:llm_judge}. The results are reported in Table~\ref{tab:ablation}.

\begin{table}[!htb]
\centering
\caption{Ablation study on report generation quality, conducted using the eRisk 2018 dataset. Average LLM-as-a-judge scores (1-5). Higher values indicate better performance.}
\label{tab:ablation}
\fontsize{7pt}{8pt}\selectfont
\setlength{\tabcolsep}{6pt}
\renewcommand{\arraystretch}{1.25}
\begin{tabular}{lccccc}
\toprule
\textbf{Variant} 
& \makecell{\textbf{Trajectory}\\\textbf{Coverage}} 
& \makecell{\textbf{Temporal}\\\textbf{Coherence}} 
& \makecell{\textbf{Sensitivity to}\\\textbf{Change Points}} 
& \makecell{\textbf{Segment-Level}\\\textbf{Specificity}} 
& \makecell{\textbf{Overall}\\\textbf{Preference}} \\
\midrule

\texttt{Traj (Full)}
& \textbf{4.9}
& \textbf{4.6}
& \textbf{4.2}
& \textbf{4.0}
& \textbf{4.1} \\

\texttt{NoSeg}
& 3.7
& 3.8
& 3.5
& 3.4
& 3.7 \\

\texttt{FixedWin}
& 3.8
& 3.6
& 3.8
& 3.5
& 3.7 \\

\texttt{NoSmooth}
& 3.9
& 3.7
& 3.7
& 3.7
& 3.6 \\

\texttt{NoStats}
& 3.9
& 3.9
& 3.8
& 3.6
& 3.8 \\

\midrule

\texttt{Base}
& 1.5
& 2.3
& 1.3
& 1.5
& 1.5 \\

\bottomrule
\end{tabular}
\end{table}

The results show a consistent pattern. First, all trajectory-aware variants outperform the \texttt{Base} configuration, confirming that modeling user history as a temporal signal significantly improves report quality. Removing segmentation (\texttt{NoSeg}) leads to a clear drop in temporal coherence and sensitivity to change points, indicating that explicitly modeling phases is crucial for capturing how mental-health signals evolve over time. 
Replacing adaptive segmentation with fixed windows (\texttt{FixedWin}) slightly improves structure but does not align well with real change points, resulting in lower coherence. Similarly, removing smoothing (\texttt{NoSmooth}) introduces more noise in the trajectory, reducing the stability of the identified phases. 
Finally, removing segment-level statistics (\texttt{NoStats}) has a smaller but noticeable impact, suggesting that simple numerical descriptors help guide the LLM in producing more precise and structured summaries.
Overall, the best performance is achieved when all components are combined, showing that trajectory construction, smoothing, adaptive segmentation, and structured prompting contribute jointly to generating coherent and interpretable reports.

\section{Limitations and Future Work}
\label{sec:discussion}

This study demonstrates that modeling mental-health signals as temporal trajectories, enriched with multidimensional classification and topic analysis, provides significant benefits for generating structured and interpretable reports. However, some limitations should be acknowledged, which also point to directions for future work. 

First, the current approach relies primarily on English data and models. Although multilingual extensions are feasible (e.g., using mBERT), performance may degrade for underrepresented languages or dialects due to limited training data and cultural differences in the expression of mental states. A second limitation concerns variability in data availability across users. Sparse or irregularly distributed digital traces may lead to incomplete or noisy trajectories, thereby reducing the reliability of temporal modeling and change-point detection. Another  limitation involves the use of large language models for report generation. While these models enhance interpretability, they may introduce hallucinations or overgeneralizations, particularly when contextual evidence is limited or ambiguous. Although a RAG component could help mitigate this issue, ensuring factual consistency and traceability remains a critical challenge.
From an ethical perspective, the system is intended as a decision-support tool rather than a diagnostic instrument. Nevertheless, risks related to privacy, data misuse, and misinterpretation of results must be carefully managed, especially when dealing with sensitive personal data.

Future work will address these limitations and explore several directions, including the integration of multimodal data (e.g., images, audio, and wearable signals), the evaluation on clinical datasets, and the incorporation of external contextual information, such as clinical annotations, to improve robustness and interpretability.

\section{Conclusions}
\label{sec:conclusions}

In this paper, we introduced an explainable framework for detecting and analyzing depression-related status shifts in user digital traces. Unlike traditional approaches that focus on isolated posts, the proposed method models mental-health signals as temporal trajectories, capturing how depression-related signals evolve over time. The framework integrates multidimensional BERT-based classification, daily aggregation and smoothing, change-point detection through piecewise linear segmentation, and LLM-based report generation to produce interpretable, phase-oriented summaries.

The experimental evaluation, conducted on Reddit and Twitter datasets, shows that incorporating temporal structure significantly improves the quality of the generated reports. In particular, trajectory-aware summaries provide better coverage of the user’s history, stronger temporal coherence, and higher sensitivity to meaningful change points compared to direct LLM-based summarization. The ablation study further confirms that each component of the pipeline (especially segmentation and temporal modeling) contributes to the overall effectiveness of the system.

Overall, the proposed framework offers a step toward more transparent and longitudinal analysis of mental-health–related signals in digital environments. While not intended for diagnostic purposes, it provides a structured and interpretable representation of behavioral evolution that can support research and assist domain experts in understanding complex temporal patterns.


\bibliographystyle{unsrtnat}
\bibliography{sn-bibliography}

\end{document}